\documentclass{article}

\usepackage{PRIMEarxiv}

\usepackage[utf8]{inputenc} 
\usepackage[T1]{fontenc}    
\usepackage[hypertexnames=false]{hyperref}       
\usepackage{url}            
\usepackage{booktabs}       
\usepackage{amsfonts}       
\usepackage{amsmath}        
\usepackage{amsthm}         
\theoremstyle{definition}

\usepackage{nicefrac}       
\usepackage{silence}        
\usepackage{microtype}      
\usepackage{fancyhdr}       
\usepackage{graphicx}       
\usepackage{algorithm}      
\usepackage{algpseudocode}  
\usepackage{listings}       
\usepackage{multirow}       
\usepackage{array}          
\graphicspath{{media/}{figures/}}     
\lstdefinelanguage{yaml}{
  keywords={true,false,null,y,n},
  sensitive=false,
  comment=[l]{\#},
  morestring=[b]',
  morestring=[b]",
}
\title{OntoKG: Ontology-Oriented Knowledge Graph Construction with Intrinsic-Relational Routing\\[0.8em]
}

\author{
Yitao Li \\
ProRata.ai \\
\texttt{yitao@prorata.ai}
\thanks{These authors contributed equally to this work}
\and
Zhanlin Liu \\
ProRata.ai \\
\texttt{zhanlin@prorata.ai} \footnotemark[1]
\and
Anuranjan Pandey \\
ProRata.ai \\
\texttt{anuranjan@prorata.ai}
\and 
Muni Srikanth \\
ProRata.ai \\
\texttt{srikanth@prorata.ai}
}

\begin{document}
\maketitle
\let\thefootnote\relax\footnotetext{Code and schema available at \url{https://github.com/Prorata-ai/OntoKG}}

\begin{abstract}
Organizing a large-scale knowledge graph into a typed property graph requires structural decisions---which entities become nodes, which properties become edges, and what schema governs these choices. Existing approaches embed these decisions in pipeline code (YAGO, DBpedia) or extract relations ad hoc (LLM-based pipelines), producing schemas that are tightly coupled to their construction process and difficult to reuse for downstream ontology-level tasks. We present an ontology-oriented approach to knowledge graph construction: one in which the schema is designed from the outset for ontology analysis, entity disambiguation, domain customization, and LLM-guided extraction---not merely as a byproduct of graph building. The core mechanism is \emph{intrinsic-relational routing}, which classifies every property as either \emph{intrinsic} (a node attribute for lookup, e.g., birth date) or \emph{relational} (a traversable graph edge, e.g., employer), and routes it to the corresponding schema module. This routing produces a declarative schema that is portable across storage backends and independently reusable.

We instantiate the approach on the January 2026 Wikidata dump (about 100\,M items). A rule-based cleaning stage identifies a 34.6\,M-entity core set from the full dump, followed by iterative intrinsic-relational routing that assigns each property to one of 94 modules (56 intrinsic, 38 relational) organized into 8 categories. With tool-augmented LLM support and human review, the schema reaches 93.3\% category coverage on the core set and 98.0\% module assignment among classified entities. Exporting this schema yields a property graph with 34.0\,M nodes and 61.2\,M edges across 38 relationship types. We validate the ontology-oriented claim through five applications that consume the schema independently of the construction pipeline: ontology structure analysis, benchmark annotation auditing, entity disambiguation (+2.4-point macro gain over YAGO~4.5 on the controlled-candidate subset of BLINK), domain customization, and LLM-guided extraction.

\end{abstract}


\section{Introduction}\label{sec:intro}

Knowledge graphs have become a foundational structure for organizing information, from domain-specific applications to world-scale knowledge bases. Among these, Wikidata stands as one of the largest collaborative knowledge graphs, containing over 100 million entities contributed by a global community of editors~\cite{vrandecic2014}. However, Wikidata's open editing model, while enabling broad coverage, comes at a cost: entity types are unconstrained, property usage is inconsistent, and no relationship schema is enforced. The result is a raw, semi-structured knowledge graph that is rich in content but difficult to query, analyze, or integrate into downstream applications at scale.

Several projects have sought to impose structure on Wikidata. DBpedia~\cite{lehmann2015} and YAGO~4.5~\cite{suchanek2024} extract and organize Wikidata content into cleaner ontological frameworks. However, these efforts focus on ontology alignment---mapping Wikidata into existing type hierarchies---rather than producing a schema that is itself designed for ontology-level applications such as structural analysis, entity disambiguation, or domain-specific subgraph extraction. Meanwhile, LLM-constructed knowledge graphs face the same disorder: recent pipelines extract entities and relations from unstructured text at scale, but the resulting graphs have ad hoc entity types, unconstrained relation vocabularies, and no shared schema governing what is extracted or how it is organized.

What is missing is an \emph{ontology-oriented} approach: one in which the schema is constructed with ontology applications in mind, so that the resulting artifact is not merely a side effect of graph building but a first-class resource for downstream tasks. Such a schema must be comprehensive in coverage, modular enough to support domain customization, and portable across storage backends---properties that existing pipeline-coupled schemas lack.

\begin{figure}[t]
\centering
\includegraphics[width=\textwidth]{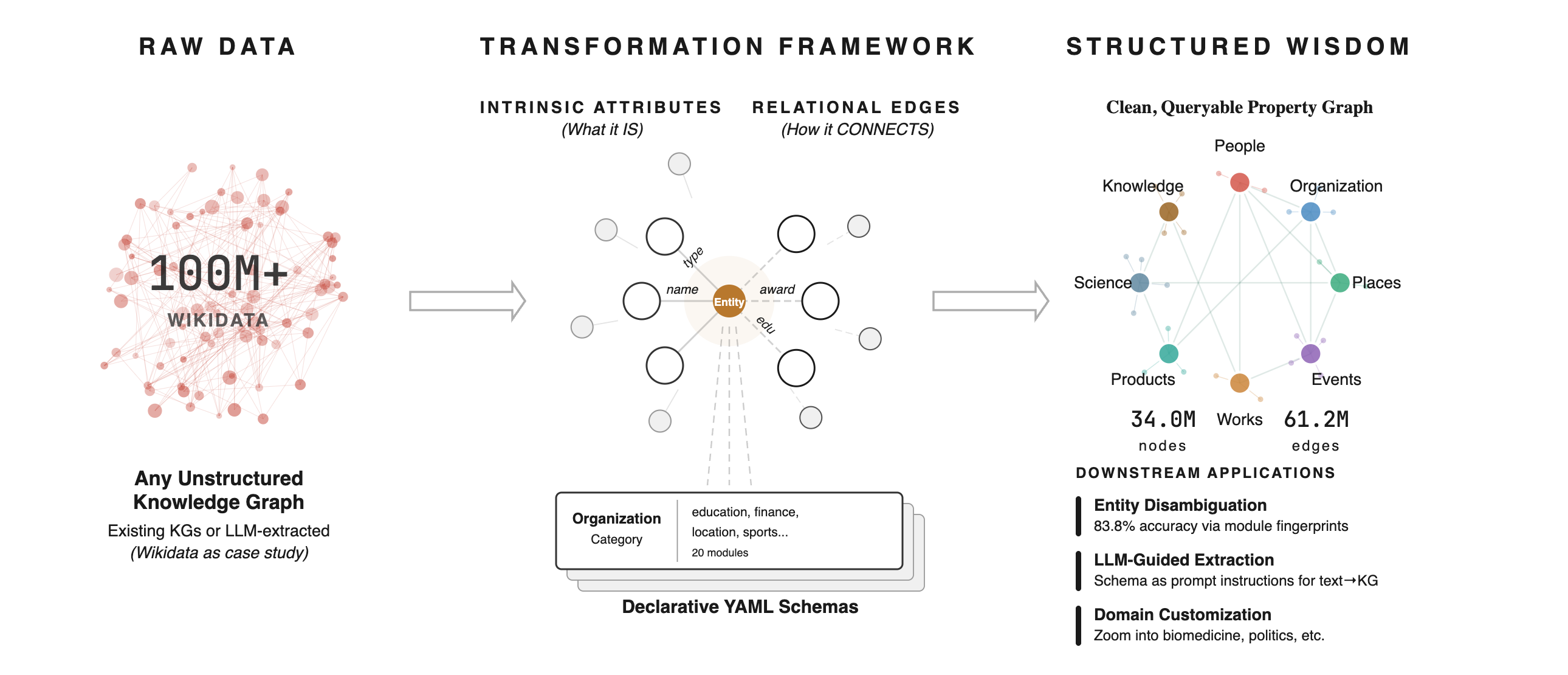}
\caption{Overview of the schema-centered knowledge graph ecosystem. A raw knowledge graph (left) is transformed through an intrinsic-relational classification framework into a typed property graph (right). The declarative schema (center) serves as the core artifact, enabling downstream applications including entity disambiguation, domain customization, and LLM-guided extraction.}
\label{fig:overview}
\end{figure}

This paper operationalizes this approach through \emph{intrinsic-relational routing}, a methodology that classifies every property as either intrinsic or relational and routes it to the corresponding schema module (Figure~\ref{fig:overview}). \emph{Intrinsic modules} route entity properties to typed tabular attributes suitable for node lookup, while \emph{relational modules} route properties to graph edges that capture traversable connections. Together, they define the \emph{node scope} (which entities appear as nodes and what properties attach to them) and the \emph{edge boundary} (which properties become traversable edges versus scalar attributes). The resulting classification does more than determine graph structure: it also organizes each entity's raw properties into semantically meaningful modules that support inspection and downstream reuse. The construction process preserves analyzable conceptual structure---explicit categories, reusable modules, principled property routing---while the produced artifact remains a declarative schema rather than a formal OWL/RDF ontology. We demonstrate the methodology on Wikidata, where the data cleaning stage (Section~\ref{sec:bulk}) identifies approximately two-thirds of the 100\,M entities as automated bulk imports---motivating a data cleaning stage before classification.

Our contributions are:

\begin{enumerate}
    \item An ontology-oriented approach for organizing a raw, open-domain knowledge graph into a typed property graph designed for downstream ontology-based applications.
    \item \emph{Intrinsic-relational routing}, a new procedure that constructs a schema and its typed property graph by distinguishing intrinsic attributes from relational connections and routing them to appropriate tabular fields or graph edges.
    \item An agentic LLM workflow for iterative schema refinement, where the LLM acts as a schema designer equipped with grounding tools that verify each identifier used in schema decisions against the knowledge graph.
    \item A Wikidata case study producing a property graph of 34.0\,M nodes and 61.2\,M edges with a portable schema of 8 categories and 94 modules, implemented in both Python and Rust.
    \item Five downstream applications demonstrating the schema as a first-class artifact: ontology analysis and topic-centered subgraph extraction, benchmark annotation auditing, entity disambiguation (outperforming YAGO~4.5 by 2.4 points on the controlled-candidate subset of BLINK), domain customization, and LLM-guided entity extraction.
\end{enumerate}

\section{Related Work}\label{sec:related}

We situate our contributions within four areas: knowledge graph entity typing, ontology modularization, LLM-based knowledge graph construction, and property graph construction.

\subsection{Entity Typing in Knowledge Graphs}

Existing entity typing systems assign entities to hierarchical type structures. DBpedia~\cite{lehmann2015} uses an ontology derived from Wikipedia infobox templates. The YAGO lineage~\cite{suchanek2007,suchanek2024} has evolved from WordNet-based typing to YAGO~4.5, which constructs a large-scale typed knowledge graph from Wikidata using schema.org as its type vocabulary. These systems primarily emphasize single-hierarchy classifications: an entity is assigned to one dominant type path.

Wikidata-specific entity typing has received growing attention. NECKAr~\cite{geiss2018} classifies Wikidata entities into three coarse classes (person, location, organization) using P31 values and graph structure, covering 8 million entities. Zhou et al.~\cite{zhou2025} diagnose inconsistencies in Wikidata's P31/P279 hierarchy. Our framework differs in two ways. First, it introduces a second classification dimension: alongside category assignment, \emph{relational modules} capture cross-cutting domain connections (military, finance, education) that span the primary category decomposition. Second, execution is deterministic and traceable given a fixed schema---each assignment follows from specific property values in the schema, requiring no learned models.

\subsection{Ontology Modularization and Faceted Classification}

The theoretical foundations for our two-layer framework draw from three traditions. Guizzardi's Unified Foundational Ontology (UFO)~\cite{guizzardi2005,guizzardi2022} distinguishes \emph{rigid sortals} (types essential to an entity's identity) from \emph{roles} and \emph{mixins} (contingent, relationally dependent types). Kiczales et al.~\cite{kiczales1997} identify an analogous pattern in software engineering: \emph{cross-cutting concerns} that cannot be cleanly encapsulated in a single module of a system's primary decomposition. Ranganathan's faceted classification~\cite{ranganathan1933} decomposes subjects into independent facets rather than forcing them into a single hierarchy; Stuckenschmidt et al.~\cite{stuckenschmidt2009} formalize ontology modularization along similar lines.

Our intrinsic-relational distinction operationalizes these ideas at scale: intrinsic modules behave like rigid sortals or primary facets (identity-providing, category-confined), while relational modules behave like mixins or cross-cutting concerns (spanning the primary category decomposition). A dominant heuristic---whether a module appears in multiple categories---guides this classification, reducing the degree of philosophical judgment that foundational ontology traditionally requires.

\subsection{LLM-Based Knowledge Graph Construction}

Recent work has applied LLMs to automate knowledge graph construction from unstructured text. Zhang and Soh~\cite{zhang2024edc} propose an extract-define-canonicalize pipeline where LLMs extract entities and relations, define their types, and canonicalize them into a coherent graph. Feng et al.~\cite{feng2024ontology} ground LLM-based KG construction in Wikidata's schema by generating competency questions, extracting relations, and matching them to Wikidata properties.

These approaches use LLMs to \emph{build} schemas from text---discovering entity types and relations during extraction. Our work takes a complementary approach: we use a pre-built, iteratively refined schema to \emph{guide} extraction (Section~5.4), providing the LLM with a grounded type taxonomy and tag vocabulary rather than asking it to invent one.

A further distinction is the role of the LLM itself. In existing pipelines, the LLM serves as an extractor---consuming text and producing graph triples. In our agentic oracle workflow (Section~\ref{sec:agentic}), the LLM acts as a \emph{schema designer}: it investigates unclassified entity types, proposes category and module assignments, and edits the declarative schema files. Grounding tools verify each identifier used in schema decisions against the knowledge graph at decision time, addressing the hallucination problem at its source rather than through post-hoc validation.

\subsection{Property Graph Construction}

Several systems address Wikidata processing at scale: the Wikidata Toolkit~\cite{erxleben2014} provides RDF exports via Java, Fernandez et al.~\cite{fernandez2013} introduce HDT for compressed RDF querying, and Nguyen and Takeda~\cite{nguyen2022} build Wikidata-lite using LMDB for offline knowledge extraction. On the formal side, Angles~\cite{angles2018} defines the property graph data model and Angles et al.~\cite{angles2020} prove information-preserving mappings from RDF to property graphs. Hogan et al.~\cite{hogan2021} identify hybrid storage---splitting data between graph and tabular engines---as an open challenge.

Our pipeline differs from these in three ways. First, we operate on Wikidata's native JSON dump rather than requiring an RDF processing stack. Second, we perform semantic classification before graph construction, so node labels and edge types are determined by the schema rather than by source structure. Third, the intrinsic-relational framework directly addresses hybrid storage: intrinsic modules produce node attributes for tabular lookup while relational modules produce typed edges for graph traversal, splitting data by access pattern from a single classification pass.

\section{Methodology}

This section presents the generic methodology for transforming a raw knowledge graph into a typed property graph with an explicit, reusable schema. The process applies to any knowledge graph whose entities carry class or type assertions and whose facts are organized as entity-attribute-value triples or entity-relationship-entity triples.

\subsection{Declarative Classification Schema}

The schema serves two purposes: it defines \emph{how to classify} entities and \emph{what to extract} from them. We represent it as a set of declarative configuration files---one per category---that jointly function as the single source of truth for both classification rules and output structure. Let $\mathcal{K} = (E, P, V)$ denote a knowledge graph where $E$ is the set of entities, $P$ the set of properties, and $V$ the set of values (which may themselves be entities).
\paragraph{Categories and gate matching.}
The schema partitions entities into $k$ categories $\{C_1, \ldots, C_k\}$. Each category $C_i$ is defined by a set of \emph{gate values} $G_i \subset V$---entity types that trigger membership. An entity $e \in E$ is assigned to category $C_i$ if any of its type-assertion properties (e.g., instance-of or subclass-of) matches a value in $G_i$:
\[
\texttt{category}(e) = C_i \quad \text{if} \quad \exists\, p \in P_{\text{type}},\; v \in V(e, p) \;\text{s.t.}\; v \in G_i
\]
where $P_{\text{type}}$ is the set of type-assertion properties and $V(e, p)$ denotes the values of property $p$ for entity $e$. Categories are evaluated in a fixed priority order; the first match wins, ensuring mutual exclusivity:
\[
\forall\, e \in E,\; |\{C_i : \texttt{category}(e) = C_i\}| \leq 1
\qquad \text{and} \qquad
C_i \cap C_j = \emptyset \;\; \forall\, i \neq j
\]
That is, the assigned categories form a partition over classified entities: each classified entity belongs to exactly one category.

\paragraph{Modules.}
Within each category $C_i$, the schema defines a set of named \emph{modules} $\mathcal{M}_i$ that group semantically related properties. Each module $m \in \mathcal{M}_i$ has three fields:

\begin{itemize}
    \item \textbf{Type} $\tau(m) \in \{\textit{intrinsic}, \textit{relational}\}$: determines whether the module's properties are routed to node attributes (for lookup) or graph edges (for traversal). This is the \emph{edge boundary} decision, made explicitly at schema design time.
    \item \textbf{Indicators} $I(m)$: a set of property-value conditions that trigger module assignment. Each indicator is a pair $(p, S_p)$ where $p \in P$ and $S_p \subseteq V$. An entity $e$ matches module $m$ if any indicator is satisfied:
    \[
    \texttt{match}(e, m) = \bigvee_{(p, S_p) \in I(m)} \begin{cases} V(e, p) \neq \emptyset & \text{if } S_p = \emptyset \\ V(e, p) \cap S_p \neq \emptyset & \text{otherwise} \end{cases}
    \]
    When $S_p = \emptyset$, the indicator is \emph{presence-based}: the property exists on the entity, regardless of its values. When $S_p \neq \emptyset$, the indicator is \emph{value-based}: the property must contain a value from the specified set.
    \item \textbf{Value properties} $\Pi(m) \subset P$: the properties extracted for matched entities under this module. For each matched entity $e$ and module $m$, the output is:
    \[
    \texttt{output}(e, m) = \{(p, v) \mid p \in \Pi(m),\; v \in V(e, p)\}
    \]
    If $\tau(m) = \textit{intrinsic}$, each $(p, v)$ pair becomes a node attribute; if $\tau(m) = \textit{relational}$ and $v \in E$, it becomes a typed edge $e \xrightarrow{m} v$ in the property graph.
\end{itemize}

An entity receives one category, zero or more intrinsic modules, and zero or more relational modules. Intrinsic modules define what an entity \emph{is}---identity-providing attributes that are typically category-confined (e.g., chemical formula for compounds, birth date for people). Relational modules define what an entity \emph{connects to}---cross-cutting domains that often span multiple categories (e.g., \texttt{award}, \texttt{military}, \texttt{religion}). The intrinsic--relational distinction is the default organizing principle; the schema designer may override it via the type assignment $\tau(m)$, for example reclassifying a module as relational when its values are entities worth traversing rather than scalar attributes. Taken together, a schema is formally defined as $\mathcal{S} = \{(C_i, G_i, \mathcal{M}_i)\}_{i=1}^{k}$, where each module $m \in \mathcal{M}_i$ carries $(\tau(m), I(m), \Pi(m))$.

\paragraph{Illustrative example.}
Consider Apple Inc.\ (Q312), whose \texttt{instance-of} (P31) values include \emph{enterprise} (Q6881511) and \emph{business} (Q4830453). The gate set $G_{\text{organizations}}$ contains Q6881511, so:
\[
\exists\, p \in P_{\text{type}},\; v \in V(\text{Q312}, p) \;\text{s.t.}\; v \in G_{\text{organizations}}
\quad\Longrightarrow\quad
\texttt{category}(\text{Q312}) = C_{\text{organizations}}
\]
The first category match wins; Apple's remaining P31 values do not trigger any other category.

Once classified, module indicators are evaluated. The \texttt{corporation} module ($\tau = \textit{intrinsic}$) activates because Apple's P31 values match its value-based indicator set and it possesses P169 (chief executive officer) and P1128 (employees). Its value properties---P1128 (employees), P414 (stock exchange)---become node attributes for lookup; for instance, P1128 records 164{,}000 employees (as of 2022) and P414 captures listing on Nasdaq with ticker symbol AAPL. The \texttt{affiliation} module ($\tau = \textit{relational}$) activates on properties like P112 (founded by) and P355 (child organization). Because $\tau(\texttt{affiliation}) = \textit{relational}$ and the values are entities, each becomes a typed edge: e.g., $\text{Q312} \xrightarrow{\texttt{affiliation}} \text{Q1961036}$ (Beats Electronics). In total, Apple receives 1 intrinsic module and 8 relational modules, illustrating how a single entity acquires both identity-defining attributes and cross-cutting graph connections.

\subsection{Iterative Schema Refinement}

The schema is not designed in a single pass. It emerges through an iterative refinement loop that alternates between classification and failure analysis. The process begins with a \emph{seed schema} $\mathcal{S}_0$---a small set of gate values, a handful of modules per category, and an initial list of value properties---that reflects the builder's intent for the target graph. Each round identifies two failure sets: \emph{unclassified entities} $E_{\emptyset}$ (no gate match) and \emph{no-module entities} $E_{\neg m}$ (classified but no module activated), then applies three decision oracles to resolve them.

\begin{algorithm}[t]
\caption{Iterative Schema Refinement}\label{alg:refinement}
\begin{algorithmic}[1]
\Require $\mathcal{K}=(E,P,V)$, seed schema $\mathcal{S}_0$, oracles $\delta_c: V \to C_i$, $\delta_m: V \to m$, thresholds $\theta_c,\theta_m$
\Ensure Refined schema $\mathcal{S}^*$
\State $\mathcal{S} \gets \mathcal{S}_0$
\State $\forall\, e \in E$: compute $\texttt{category}(e)$; $\forall\, m \in \mathcal{M}_{\texttt{category}(e)}$: compute $\texttt{match}(e,m)$ \Comment{initial classification}
\State $E_{\emptyset} \gets \{e : \texttt{category}(e) = \emptyset\}$, \quad $E_{\neg m} \gets \{e : \texttt{category}(e) = C_i \land \forall\, m \in \mathcal{M}_i,\,\neg\texttt{match}(e,m)\}$
\State $r_c \gets 1 - |E_{\emptyset}|/|E|$
\State $N_c \gets |E|-|E_{\emptyset}|$; \quad $r_m \gets \begin{cases}0, & N_c=0 \\ 1-|E_{\neg m}|/N_c, & N_c>0\end{cases}$
\While{$r_c < \theta_c$ \textbf{or} $r_m < \theta_m$}
    \State $T \gets \textsc{CandidateTypes}(E_{\emptyset})$ \Comment{gate expansion}
    \ForAll{$t \in T$}
        \State $G_{\delta_c(t)} \gets G_{\delta_c(t)} \cup \{t\}$
    \EndFor
    \ForAll{categories $C_i$} \Comment{gate--module sync}
        \ForAll{new $g \in G_i$ s.t.\ $\not\exists\, m \in \mathcal{M}_i,\,(p, S_p) \in I(m): g \in S_p$}
            \State $m \gets \delta_m(g)$; select $(p,S_p)\in I(m)$ with $S_p\neq\emptyset$ (or create one); $S_p \gets S_p \cup \{g\}$
        \EndFor
    \EndFor
    \State Update or create $m \in \mathcal{M}_i$ for each $i$: $I(m),\Pi(m),\tau(m)$ \Comment{module refinement}
    \State $\forall\, e \in E$: compute $\texttt{category}(e)$; $\forall\, m \in \mathcal{M}_{\texttt{category}(e)}$: compute $\texttt{match}(e,m)$ \Comment{reclassify}
    \State $E_{\emptyset} \gets \{e : \texttt{category}(e) = \emptyset\}$, \quad $E_{\neg m} \gets \{e : \texttt{category}(e) = C_i \land \forall\, m \in \mathcal{M}_i,\,\neg\texttt{match}(e,m)\}$
    \State $r_c \gets 1 - |E_{\emptyset}|/|E|$
    \State $N_c \gets |E|-|E_{\emptyset}|$; \quad $r_m \gets \begin{cases}0, & N_c=0 \\ 1-|E_{\neg m}|/N_c, & N_c>0\end{cases}$
\EndWhile
\State \Return $\mathcal{S}^* \gets \mathcal{S}$
\end{algorithmic}
\end{algorithm}

Algorithm~\ref{alg:refinement} formalizes the loop. The $\textsc{CandidateTypes}$ function (line 6) selects type values from $E_{\emptyset}$ for gate expansion using two criteria: \emph{high frequency} (types with the most unclassified instances, maximizing coverage gain) and \emph{high reference count} (types whose entities are heavily referenced by already-classified entities, maximizing graph connectivity even when instance counts are low). In gate--module synchronization, if multiple indicators in $I(m)$ are eligible, we reuse one on a type-assertion property; if none exists, we create a new value-based indicator. The computational steps---classification, failure partitioning, rate computation---are deterministic and parallelizable. The bottleneck lies in three decision oracles that require semantic judgment:

\begin{enumerate}
    \item \textbf{Category oracle} $\delta_c: V \to C_i$\textbf{.} Given an unclassified type value, assign it to a category. This drives the gate expansion step.
    \item \textbf{Module oracle} $\delta_m: V \to m \in \mathcal{M}_i$\textbf{.} Given a new gate value with no matching indicator, assign it to a module within its category. This enforces the gate--module synchronization invariant $\forall\, g \in G_i,\; \exists\,(p, S_p)\in I(m)\ \text{s.t.}\ g \in S_p$.
    \item \textbf{Refinement oracle.} Given the current schema and failure analysis, decide whether to create, merge, or split modules, and set their parameters $I(m)$, $\Pi(m)$, $\tau(m)$ in the module refinement step.
\end{enumerate}

Traditionally, these oracles require expensive human annotation or purpose-built classifiers trained on labeled data. In Section~4, we describe an alternative: an agentic LLM workflow equipped with grounding tools (entity queries, property lookups, sample inspection) that implements all three oracles, reducing both human effort and hallucination risk.

\paragraph{Convergence.} The schema evolves through additive updates to gates and indicators, with periodic module restructuring during refinement. Under fixed category priority, gate expansion typically improves category coverage, so $r_c$ is non-decreasing in most rounds. By contrast, $r_m$ is not guaranteed to be monotonic: reassignment and module updates can temporarily increase or decrease $|E_{\neg m}|$ before later rounds recover coverage. The loop terminates when both rates reach their targets or when no further oracle assignments are made. In practice, we set $\theta_c$ and $\theta_m$ to about $0.9$, since a long tail of rare or ambiguous types may not belong in any category; the oracles correctly decline to assign them, and the remaining gap reflects genuine coverage limits of the schema rather than algorithmic failure.

\section{Wikidata Case Study}

We apply the methodology from Section~3 to Wikidata (January 2026 dump), which contains approximately 100\,M entities. We resolve labels and descriptions using a fixed language fallback order: \texttt{en} $\rightarrow$ \texttt{en-us} $\rightarrow$ \texttt{en-gb} $\rightarrow$ \texttt{mul}. The pipeline proceeds through data cleaning, schema instantiation and refinement, and output generation, with an agentic LLM workflow implementing the decision oracles.

\subsection{Data Cleaning}\label{sec:bulk}

The input is a single Wikidata JSON dump (100\,GB compressed), which we partition into 24 independent Zstandard-compressed JSON Lines files (\texttt{.jsonl.zst}), each containing approximately 4.5\,M entities, enabling parallel processing across all downstream stages.

The first step establishes \emph{graph scope}: which entities belong in the graph at all. Wikidata's 100\,M entities include a substantial proportion of automated bulk imports and infrastructure entities that are not suitable for an encyclopedic knowledge graph. Data cleaning separates these from genuine editorial content through a priority-ordered cascade:

\begin{enumerate}
    \item \textbf{Structural classification.} P31-based detection using 45 infrastructure QIDs (e.g., Wikimedia disambiguation pages, categories) and 28 bulk-import QIDs (e.g., scientific article, gene) identifies entities that are structurally outside the encyclopedic scope.
    \item \textbf{Source signature matching.} 19 known import databases (e.g., CrossRef, astronomy catalogs, taxonomy databases) are detected via characteristic property combinations that fingerprint their bulk-imported entities.
    \item \textbf{Curation score.} A weighted curation score\footnote{36 properties across five tiers weighted 1--3: strong signals (images, coordinates, website) score 3; medium signals (awards, notable work) score 2; weak signals (country, industry) score 1; with separate people-specific tiers for occupation, education, etc.} protects notable entities from false positives---an entity with rich editorial signals is retained even if its P31 type matches a bulk-import pattern.
    \item \textbf{Ratio-based safety net.} Remaining entity types with a bulk-import property ratio $\geq 70\%$ and insufficient curation signals (score $< 3$) are automatically classified as non-core.
\end{enumerate}

The cleaning stage is rule-based and deterministic given fixed rule definitions: every tier assignment traces to a specific match. Prior work on Wikidata quality focuses on statement-level accuracy and completeness~\cite{piscopo2019,shenoy2021,paulheim2017}; our cleaning addresses a complementary dimension---entity-level provenance---and is designed to separate likely human-curated content from likely automated bulk imports~\cite{steiner2014,pellissier2016}. After cleaning, 34.6\,M entities remain as the \emph{core dataset} for classification.

\subsection{Schema Instantiation and Refinement}

The seed schema $\mathcal{S}_0$ was constructed by incorporating three sources: Schema.org's type hierarchy for broad category structure, YAGO 4.5's curated taxonomy for entity type coverage, and Wikidata's own EntitySchema definitions for property-level detail. From these, we derived eight categories: \emph{people}, \emph{places}, \emph{creative\_works\_media}, \emph{knowledge}, \emph{science}, \emph{organizations}, \emph{events\_actions}, and \emph{products\_artifacts}. Gate matching checks P31 (instance of) and P279 (subclass of) values against curated QID sets in a fixed priority order. The schema is defined in eight YAML files, each specifying gate QIDs, core properties, and module definitions with indicators and value properties.

\paragraph{Refinement process.}
We applied Algorithm~\ref{alg:refinement} iteratively over the core dataset. The Rust classifier processes the full 34.6\,M entity set in approximately five minutes ($\sim$110{,}000 entities/second), making A/B comparison practical for every schema edit. All runtime measurements in this section were recorded on a local MacBook Pro with Apple M4 Max (16 CPU cores, 48\,GB RAM). Each round applied the three oracles (Section~\ref{sec:agentic} details the implementation): $\delta_c$ expanded gate sets with unclassified P31 types, $\delta_m$ synchronized new gate values to module indicators, and the refinement oracle merged, split, or created modules as needed.

\begin{figure}[t]
\centering
\includegraphics[width=\textwidth]{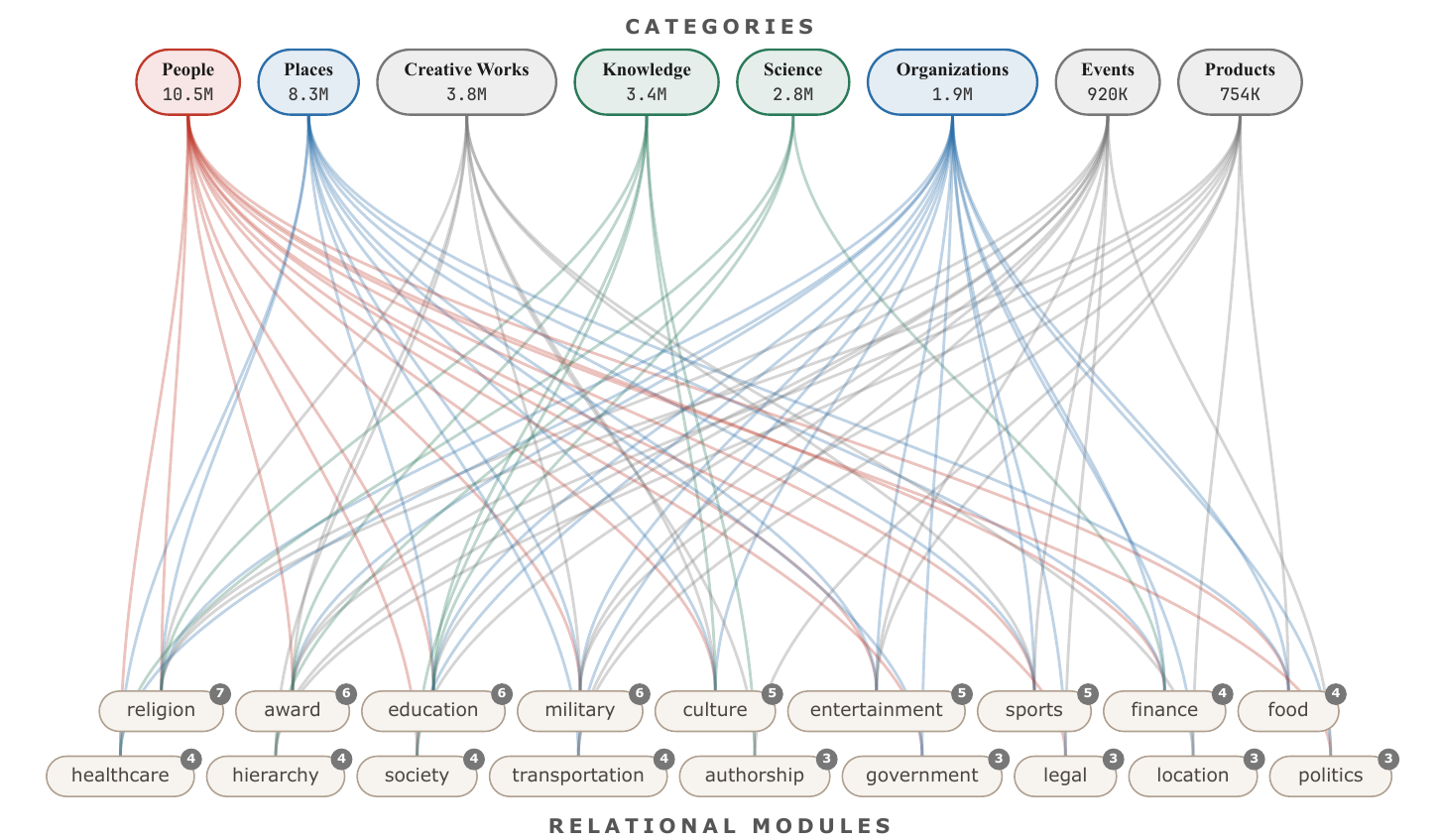}
\caption{Bipartite view of 8 categories (top) connected to 18 cross-category relational modules with span $\geq 3$ (bottom). Badges indicate category span. Two additional modules (\texttt{affiliation} and \texttt{technology}, each spanning 2 categories) are omitted for clarity. The full schema comprises 56 intrinsic modules, 18 single-category relational modules, and 20 cross-category relational modules.}
\label{fig:bipartite}
\end{figure}

\paragraph{Final schema.}
After convergence, the refined schema $\mathcal{S}^*$ achieves a classification rate of 93.3\% (32.3\,M of 34.6\,M core entities) and a module-assignment rate of 98.0\% among classified entities (no-module rate reduced from 15.8\% to 2.0\%). The schema comprises 94 modules: 56 intrinsic and 38 relational. Intrinsic modules are category-confined (e.g., \texttt{film} in creative\_works\_media, \texttt{river} in places). Relational modules capture cross-cutting domains: 18 span three or more categories (e.g., \texttt{military} in 6, \texttt{religion} in 7) and 2 span exactly two (\texttt{affiliation} and \texttt{technology}), while 18 single-category modules are typed relational because their value properties create entity-to-entity edges (e.g., \texttt{family} in people, \texttt{genomics} in science). Figure~\ref{fig:bipartite} visualizes the complete module landscape.

\subsection{Agentic Oracle Implementation}\label{sec:agentic}

The three decision oracles ($\delta_c$, $\delta_m$, and the refinement oracle) are implemented as an agentic workflow powered by Claude Opus 4.6\footnote{Anthropic's Claude Opus 4.6 (claude-opus-4-6), accessed via Claude Code CLI.}. Two challenges motivate the tool design. First, LLMs hallucinate Wikidata identifiers: given a label, an LLM may fabricate a plausible but nonexistent QID, and given a QID, it may invent an incorrect label. Second, the agent cannot reliably classify an unfamiliar entity type from its label alone---it requires seeing sample entities, instance counts, and descriptions to make an informed category decision. To address both challenges, the agent is equipped with grounding tools that provide verified information from the knowledge graph:

\begin{itemize}
    \item \textbf{LMDB label lookup} (\texttt{query\_lmdb}): resolves QIDs and PIDs to their canonical English labels, and vice versa, against a memory-mapped store\footnote{We use LMDB (Lightning Memory-Mapped Database) for its zero-copy reads and constant-time key lookups, enabling instant label resolution without loading the full database into memory.} built once from a full scan of the Wikidata dump. Every identifier the agent references is verified through this tool.
    \item \textbf{Live P31 query} (\texttt{query\_p31}): queries the Wikidata SPARQL endpoint to retrieve entity counts, labels, descriptions, and sample instances for a given P31 type---allowing the agent to inspect what a type value actually represents before deciding its category or module assignment.
    \item \textbf{Tag validator} (\texttt{tag\_validator}): validates YAML schema files for structural correctness, gate--module synchronization, QID/PID format, and cross-category gate exclusivity.
    \item \textbf{Category P31 analysis} (\texttt{analyze\_category\_p31}): produces per-category coverage statistics, P31 frequency tables, P31$\times$module cross-tabulations, and reports of entities with no module assignment---diagnosing exactly where the schema fails.
    \item \textbf{Unclassified hub analysis} (\texttt{find\_unclassified\_hubs}): identifies high-reference unclassified entities in the classified Parquet files using DuckDB, groups them by P31 type, and ranks by inbound reference count---surfacing the entity types with the greatest impact on graph connectivity.
\end{itemize}

A key design choice in the YAML schema is that every QID and PID must carry an inline comment with its canonical English label, verified against LMDB by the tag validator. This makes the schema files self-documenting: the agent can review and edit definitions by reading the YAML directly, without querying each identifier individually.

Live SPARQL queries are used only for exploratory inspection of candidate P31 types during refinement; final classification and evaluation metrics are computed from the fixed January 2026 dump.

\paragraph{Investigation phase.}
The agent examines two priority lists: the top-$k$ most frequent unclassified entity types (for coverage) and the top-$k$ highest-reference hub types (for graph connectivity---these may be infrequent but heavily referenced by classified entities). The values of $k$ and the number of refinement rounds are chosen adaptively from round to round based on observed failure patterns and review capacity, rather than fixed globally. For each candidate, the agent queries labels and sample entities via the grounding tools, then generates a report recommending which category ($\delta_c$), which module ($\delta_m$), and whether new modules are needed. This report is presented for human review; the reviewer may accept, reject, or annotate individual recommendations.

\paragraph{Implementation phase.}
The agent applies the approved changes to the YAML files, using the label lookup to ensure correct QID/PID annotations and the tag validator to check structural integrity. The human operator then triggers reclassification via the Rust classifier to measure the coverage improvement, closing the loop for the next refinement round.

\subsection{Classified Result}

The primary output of the classification pipeline is a set of Parquet files (13 columns per entity) containing category, module assignments, and claims partitioned into four buckets: core (shared identity attributes), intrinsic (category-specific scalar properties), relational (entity-to-entity connections), and unclaimed (remaining properties). Each claim carries its property label, resolved value label, qualifiers, and a pre-generated natural-language sentence (e.g., ``James Humphrey Walwyn's occupation is military officer (military branch: Royal Navy).'') with all QIDs and PIDs replaced by their English labels. These self-contained files serve as the single source for all downstream outputs: the core and intrinsic buckets provide lookup-table information for each node, the relational bucket drives graph edge construction, and the sentence field enables direct use in LLM prompts or text-based retrieval without additional ID resolution.

For property graph database import (Neo4j), a three-pass export pipeline produces graph-ready CSVs:

\begin{enumerate}
    \item \textbf{Node export.} One CSV file per category (8 files), each containing entity QID, label, description, and core properties as node attributes. Intrinsic module properties are included as additional columns on the node.
    \item \textbf{Edge export.} One CSV file per relational module (38 files), each containing source QID, target QID, relationship type (module name), and qualifier properties. Only value properties whose targets are themselves entities ($v \in E$) produce edges.
    \item \textbf{Stub node export.} Edge targets that are not in the core dataset---entities referenced by relational modules but excluded during data cleaning---are exported as stub nodes with QID and label only, preserving graph connectivity without inflating the classified node set.
\end{enumerate}

This maps directly to the intrinsic--relational distinction: intrinsic modules become node attributes (suitable for tabular stores or property lookups), while relational modules become typed edges (suitable for graph traversal). The separation ensures that each storage engine carries only the data matching its primary access pattern.

The final graph contains 32.3\,M classified entity nodes, 1.7\,M stub nodes, and 61.2\,M edges across 38 relationship types (34.0\,M total nodes). The raw CSV export totals 9.1\,GB (1.7\,GB gzipped). The pipeline is implemented in both Rust and Python: the Rust exporter processes all 24 shards in approximately 11 minutes; the Python implementation provides a single-threaded reference for validation and ad-hoc analysis.

\section{Applications}

The preceding sections constructed the schema; this section demonstrates why the construction is ontology-oriented. Each application consumes the declarative schema or the classified output independently of the original Wikidata dump and the classification pipeline: (1) ontology structure analysis and domain-specific subgraph extraction, (2) benchmark annotation auditing via ontological classification, (3) entity disambiguation via module-based type profiles, (4) domain customization through module decomposition, and (5) LLM-guided entity extraction using the schema as prompt instructions.

\subsection{Ontology Structure}

The intrinsic-relational distinction produces a natural two-level ontology that can be analyzed structurally. Figure~\ref{fig:bipartite} presents the cross-category relational structure: 8 categories (top) connected to 18 relational modules spanning 3 or more categories (bottom). The badge on each module indicates its category span; the connections between categories and modules form the schema's relational backbone.

The bipartite structure reveals two patterns. First, relational modules vary widely in span: \texttt{religion} connects 7 categories, while \texttt{authorship} and \texttt{politics} each bridge 3. Second, the modules form natural domain clusters visible in Figure~\ref{fig:bipartite}: governance-related modules (\texttt{government}, \texttt{legal}, \texttt{politics}) share the same category connections, as do cultural modules (\texttt{culture}, \texttt{entertainment}, \texttt{sports}, \texttt{religion}) and economic modules (\texttt{finance}, \texttt{food}, \texttt{transportation}). These clusters suggest that the relational modules capture coherent thematic domains rather than arbitrary property groupings.

\begin{figure}[t]
\centering
\includegraphics[width=0.75\columnwidth]{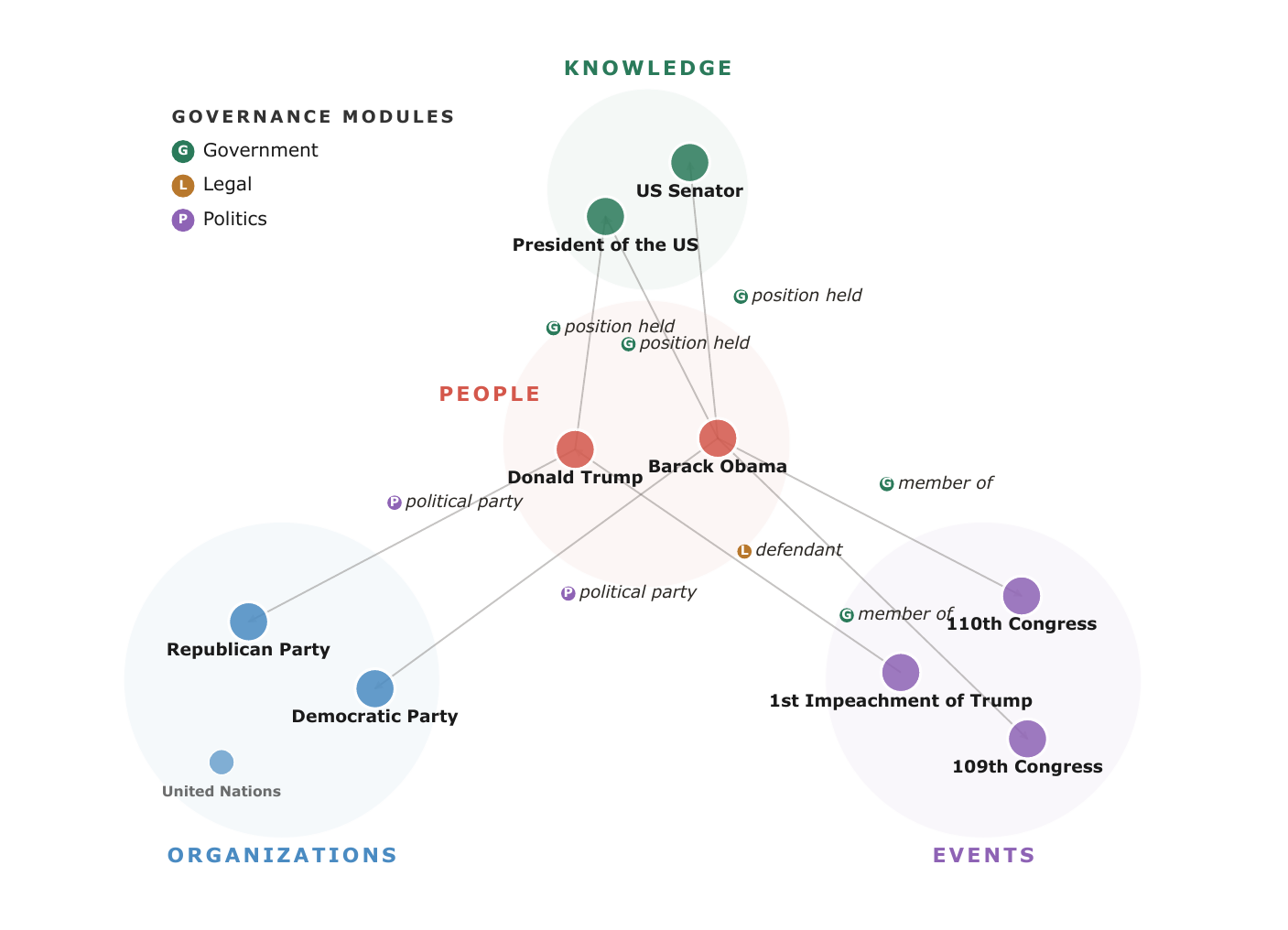}
\caption{Governance domain subgraph extracted by selecting three relational modules: \texttt{government} (green), \texttt{legal} (red), and \texttt{politics} (purple). The subgraph spans four categories---People, Knowledge, Organizations, and Events---with edge labels showing the underlying Wikidata properties (e.g., position held, political party, member of).}
\label{fig:governance-graph}
\end{figure}

These domain clusters directly enable topic-centered subgraph extraction. Selecting a cluster of relational modules produces a self-contained graph with its own node types, edge types, and schema. Figure~\ref{fig:governance-graph} demonstrates this with the governance cluster: selecting \texttt{government}, \texttt{legal}, and \texttt{politics} yields a geopolitical subgraph spanning People (Obama, Trump), Knowledge (positions like US Senator), Organizations (political parties), and Events (congressional sessions, impeachment proceedings). Similarly, \texttt{healthcare} + \texttt{education} + Science yields a biomedical research subgraph, and \texttt{culture} + \texttt{entertainment} + \texttt{sports} yields a cultural analysis subgraph---each derived entirely from module selection.

Beyond schema-level analysis, the classification system also operates at the individual entity level, transforming an unstructured bag of Wikidata properties into an organized view. Table~\ref{tab:property-org} illustrates this with a live Wikidata query for Apple Inc.\ (Q312), which currently carries 248 raw properties with no grouping beyond property IDs. The classifier routes it to \texttt{organizations} and assigns properties to structured groups: core properties (inception, headquarters, country) provide entity identity; the intrinsic module \texttt{corporation} captures corporate attributes such as employees and products for tabular lookup; and relational modules route properties to typed graph edges---\texttt{affiliation} connects Apple to its founders, owners, and subsidiaries, \texttt{finance} links to stock exchanges and executives, and \texttt{location} anchors the entity geographically. The remaining properties are predominantly external identifiers (database cross-references, social media links) that fall outside the ontological schema.

\begin{table}[t]
\centering
\caption{Property organization for Apple Inc.\ (Q312), showing selected properties from a live Wikidata query (248 total). Core and intrinsic properties map to tabular output; relational properties become typed graph edges.}
\label{tab:property-org}
\small
\begin{tabular}{@{}lllp{7.2cm}@{}}
\toprule
Type & Module & Property & Sample Values \\
\midrule
\multirow{2}{*}{}
  & \multirow{2}{*}{\texttt{core}} & P571 inception & 1976-04-01 \\
  & & P856 official website & https://apple.com/ \\
\midrule
\multirow{2}{*}{\textbf{Intrinsic}}
  & \multirow{2}{*}{\texttt{corporation}}
  & P1128 employees & 164,000; 154,000; 147,000; \ldots \\
  & & P1056 products & Apple Watch, iPad, iPhone, \ldots \\
\midrule
\multirow{10}{*}{\textbf{Relational}}
  & \multirow{5}{*}{\texttt{affiliation}}
  & P112 founded by & Steve Wozniak, Ronald Wayne, Steve Jobs \\
  & & P127 owned by & Vanguard Group, Berkshire Hathaway, BlackRock \\
  & & P355 child organization & Beats Electronics, Anobit, Braeburn Capital, \ldots \\
  & & P361 part of & Nasdaq-100, Dow Jones, Big Tech, \ldots \\
  & & P463 member of & Computer \& Communications Industry Assoc., \ldots \\
\cmidrule(l){2-4}
  & \multirow{2}{*}{\texttt{finance}}
  & P169 CEO & Tim Cook, Steve Jobs, Gil Amelio, \ldots \\
  & & P414 stock exchange & Tokyo Stock Exchange, Nasdaq \\
\cmidrule(l){2-4}
  & \multirow{3}{*}{\texttt{location}}
  & P276 location & United States \\
  & & P495 country of origin & United States \\
  & & P740 location of formation & Los Altos \\
\midrule
\multicolumn{2}{@{}l}{\textbf{Unclaimed}} & \multicolumn{2}{l}{external IDs, social media links, minor metadata} \\
\bottomrule
\end{tabular}
\end{table}

\subsection{Benchmark Annotation Auditing}
\label{sec:auditing}

Because the gate-based classifier uses only Wikidata ontological types (P31/P279) and is not trained on NER labels, it can serve as an external auditor of benchmark annotations. We evaluate it on two datasets derived from the same CoNLL-2003 Reuters corpus: AIDA-YAGO~\cite{hoffart2011}, whose mentions were manually linked to 2011 Wikipedia, and CleanCoNLL~\cite{rucker2023}, a 2023 re-annotation of the corpus with revised labels and refreshed entity links. We resolve linked Wikipedia titles to Wikidata QIDs so that each entity can be passed through the ontological classifier.

Across the full linked AIDA-YAGO and CleanCoNLL entity sets, the classifier assigns an ontological category to 95.1\% and 93.9\% of entities, respectively; among classified entities, these categories align with the benchmark NER labels for 93.2\% and 97.9\%, respectively. To enable a direct three-way comparison, we restrict attention to the 5,429 entities that appear in both datasets with the same Wikidata QID; Table~\ref{tab:auditing} summarizes the results.

\begin{table}[t]
\centering
\caption{Three-way agreement on 5,429 shared entities (matched by Wikidata QID) between AIDA-YAGO and CleanCoNLL. The YAML classifier has no NER-specific tuning; a complete case-level breakdown of disagreement categories is provided in the project repository.}
\label{tab:auditing}
\begin{tabular}{lrl}
\toprule
Scenario & Count & Interpretation \\
\midrule
All three agree & 4,840 & Consensus \\
\addlinespace
AIDA wrong, Clean + Ours agree & 180 & AIDA error confirmed \\
Clean wrong, AIDA + Ours agree & 23 & CleanCoNLL error confirmed \\
All three disagree & 4 & Different error modes \\
\addlinespace
Both NER differ from ontology & 84 & Linking errors or boundary cases \\
Unclassified (no gate match) & 298 & Cannot evaluate \\
\bottomrule
\end{tabular}
\end{table}

More importantly, when AIDA-YAGO and CleanCoNLL disagree, the ontological classifier can act as a tiebreaker. As Table~\ref{tab:auditing} shows, it sides with CleanCoNLL in 180 cases and with AIDA-YAGO in 23---a 7.8:1 ratio that supports most of CleanCoNLL's corrections while exposing a smaller set of over-corrections. In 4 cases all three systems disagree.

We now walk through the table rows to characterize each disagreement type.

\paragraph{AIDA errors confirmed (180).} The largest group consists of clear AIDA-YAGO errors that CleanCoNLL independently corrected. The dominant patterns are national sports teams tagged as LOC instead of ORG (e.g., ``Brazil'' referring to the national football team) and outdated entity links pointing to disambiguation pages rather than the intended entity.

\paragraph{CleanCoNLL errors confirmed (23).} Most are systematic LOC$\rightarrow$ORG over-corrections: the majority involve physical locations such as airports, mines, museums, ports, and government buildings that CleanCoNLL re-tagged as organizations, while the Wikidata ontology and AIDA-YAGO both treat them as locations.

\paragraph{Both NER differ from ontology (84).} This group contains two distinct phenomena. First, 30 are entity-linking errors: 22 disambiguation page links shared across both datasets, plus 8 cases where the mention is linked to the wrong entity entirely. For example, the headline ``CRAWLEY FORCED TO SIT AND WAIT'' refers to cricketer John Crawley, and both datasets correctly tag it PER, but the entity link points to the town of Crawley (Q844908), which our system classifies as a place. In these cases the NER tags are correct for the intended referent; only the entity links are wrong. Second, the remaining 54 are genuine boundary cases where NER and ontological classification apply different criteria: central banks (ORG by function vs.\ institutional places by Wikidata typing), publications (ORG vs.\ creative works), and historical countries (LOC vs.\ abstract historical concepts). These are not errors in either system.

The result is not simply that the ontology ``beats'' benchmark labels, but that it provides an independent, interpretable signal for separating annotation mistakes, linking mistakes, and task-definition differences.

\subsection{Entity Profile and Disambiguation}

The module system produces structured entity profiles that serve as discriminative type annotations for entity disambiguation. We evaluate this on the BLINK benchmark using Voyage rerank-2.5, comparing our module-based type annotations against YAGO 4.5's curated type hierarchy. To isolate the effect of type annotations from candidate retrieval, we compare on the 12,323 mentions where both systems return the same candidate entities---so observed accuracy differences are primarily attributable to type annotations under this controlled candidate set, rather than candidate coverage.

\begin{table}[t]
\centering
\caption{Entity disambiguation accuracy on BLINK benchmark (12,323 controlled mentions with identical candidate lists). Module-based type annotations vs.\ YAGO 4.5 types, using Voyage rerank-2.5.}
\label{tab:disambiguation}
\begin{tabular}{lrrrr}
\toprule
Group & Candidates & Mentions & YAGO & Module-Based (Our Approach)\\
\midrule
1 & $<5$ & 10,284 & 91.2\% & 91.6\% \\
2 & 5--9 & 1,666 & 84.4\% & 85.4\% \\
3 & 10--19 & 348 & 82.5\% & 82.8\% \\
4 & $\geq 20$ & 25 & 68.0\% & 76.0\% \\
\midrule
\textbf{Macro Avg} & & -- & \textbf{81.5\%} & \textbf{83.9\%} \\
\bottomrule
\end{tabular}
\end{table}

The module-based approach achieves a 2.4 percentage point improvement in macro average accuracy (83.9\% vs.\ 81.5\%). Gains are largest in the hardest group (20+ candidates, +8.0 points), where richer type annotations provide the most disambiguation signal; because this bucket contains only 25 mentions, the effect size should be interpreted as directional. The improvement is driven by higher annotation density: module-based candidates carry an average of 2.52 type labels per entity compared to YAGO's 1.42. Three mechanisms contribute: (1) relational modules extract fine-grained attributes (e.g., the \texttt{sports} module routes to playing position via P413, distinguishing ``running back'' from ``defensive end'' where YAGO assigns identical ``American Football Player'' types); (2) the module system recovers specific occupations for entities YAGO labels only as generic ``Person'' (30.8\% of YAGO's exclusive failures); and (3) intrinsic modules provide domain signals (e.g., \texttt{art}, \texttt{comics}) that further distinguish entities sharing identical instance-of types.

\subsection{Schema Customization}

The declarative YAML schema is designed as a parameterized schema generator: the generic definitions (8 categories, 94 modules) serve as a template that domain experts can customize without code changes. Customization follows three operations: \emph{selecting} which categories and modules to include, \emph{decomposing} broad modules into fine-grained facets, and \emph{adding} domain-specific gate QIDs or indicators. The entire pipeline (classification, export, graph construction) adapts to whatever YAML definitions are provided.

We demonstrate this with the \texttt{education} module in people, which contains approximately 90 P106 occupation indicators spanning natural sciences, engineering, humanities, social sciences, and teaching. In the generic schema, a marine biologist and a medieval historian both receive the same \texttt{education} tag with no further distinction. By decomposing this single module into 9 domain facets---e.g., \texttt{natural\_science}, \texttt{humanities}, \texttt{engineering}---users obtain fine-grained classification while preserving the same YAML structure (Appendix~\ref{app:customization} lists all 9 facets and a sample YAML definition). Only the indicator values change; the pipeline works unchanged.

This enables targeted graph construction: a biomedical researcher selects \texttt{natural\_science} + \texttt{teaching} to produce a focused graph of life scientists and medical educators, while a digital humanities project selects \texttt{humanities} + \texttt{literary} + \texttt{information} to distinguish historians, linguists, literary scholars, and librarians---all of whom are conflated under the generic \texttt{education} module.

\subsection{LLM-Guided Entity Extraction}

The YAML schema serves as a lightweight ontological structure for LLM-based entity extraction. The 8 categories define an entity taxonomy, while module names act as fine-grained semantic tags. An LLM can use this structure to extract entities from unstructured text and classify them with contextually relevant facets---without fine-tuning.

A prompt template is constructed directly from the schema: categories provide the entity type taxonomy (e.g., Person, Organization, Place, Science), and each category's module names provide the tag vocabulary (e.g., for Person: entertainment, education, politics, sports, military, finance). For readability, these display labels map directly to canonical schema categories (\texttt{people}, \texttt{organizations}, \texttt{places}, \texttt{science}, \texttt{knowledge}, \texttt{events\_actions}, \texttt{products\_artifacts}, \texttt{creative\_works\_media}). The LLM receives a text passage and returns structured output---entity name, category, and context-relevant tags (Appendix~\ref{app:prompt} provides the full prompt template). Table~\ref{tab:extraction} illustrates the output on a passage about the Mercury automobile brand.

\begin{table}[t]
\centering
\caption{LLM-guided entity extraction from a Wikipedia passage about Mercury (automobile), using the YAML schema as prompt instructions. Model: GPT-5-mini.}
\label{tab:extraction}
\begin{tabular}{lll}
\toprule
Entity & Category & Tags \\
\midrule
Mercury & Product & brand, vehicle, transportation \\
Ford Motor Company & Organization & corporation, transportation \\
Edsel Ford & Person & creativity \\
Lincoln-Mercury Division & Organization & affiliation, corporation \\
\bottomrule
\end{tabular}
\end{table}

This approach has two advantages over ad-hoc extraction schemas. First, the tag vocabulary is grounded in a schema that has been iteratively refined against 32.3\,M classified entities, supporting broad coverage with reduced redundancy in practice. Second, extracted entities can be directly linked to the knowledge graph via category and module alignment, enabling seamless integration of LLM-extracted information with the existing graph structure.

Taken together, these five applications demonstrate that the declarative schema is a reusable artifact in its own right---supporting ontology analysis, benchmark auditing, entity disambiguation, domain customization, and information extraction independently of the graph construction pipeline that produced it.

\section{Conclusion}

We presented an ontology-oriented approach to knowledge graph construction grounded in a single design principle: routing each property as either intrinsic---stored as a node attribute for lookup---or relational---exported as a traversable edge. Formalizing this distinction into a declarative schema brings two benefits. First, it makes structural decisions explicit and inspectable, replacing ad-hoc property handling embedded in construction code. Second, it decouples the schema from the pipeline that produced it, enabling independent reuse for downstream tasks such as those demonstrated in the applications section.

Several directions remain open. The current schema was refined against the January 2026 Wikidata dump, but graph expansion from external sources such as LLM-extracted knowledge graphs will introduce relation types that do not align cleanly with the existing module inventory. Accommodating such expansion requires schema-extension methods that can map new relations to existing modules when possible, extend modules when needed, and introduce new modules when no faithful alignment exists. A systematic study of schema-guided retrieval---using the intrinsic-relational structure to shape context selection for retrieval-augmented generation---is a natural next step toward closing the loop between graph construction and downstream LLM performance.

\bibliographystyle{unsrt}
\bibliography{references}

\newpage
\appendix
\section{Schema Customization: Education Module Decomposition}\label{app:customization}

Table~\ref{tab:facets} summarizes the decomposition of the generic \texttt{education} module (people category) into 9 domain-specific facets totaling 87 occupations.

\begin{table}[h]
\centering
\caption{Decomposition of the \texttt{education} module into 9 domain facets.}
\label{tab:facets}
\begin{tabular}{llr}
\toprule
Facet & Domain & Occupations \\
\midrule
natural\_science & Life sciences, taxonomy, ecology & 22 \\
humanities & History, philosophy, languages, archaeology & 16 \\
teaching & Pedagogy at all levels & 13 \\
engineering & Applied sciences, computing & 11 \\
physical\_science & Physics, chemistry, earth sciences, math & 9 \\
social\_science & Economics, sociology, psychology & 6 \\
general\_academic & Cross-domain academic roles & 5 \\
literary & Literary criticism and scholarship & 3 \\
information & Library and archival sciences & 2 \\
\midrule
\textbf{Total} & & \textbf{87} \\
\bottomrule
\end{tabular}
\end{table}

Listing~\ref{lst:yaml} shows a sample YAML definition for the \texttt{natural\_science} facet.

\begin{lstlisting}[
  basicstyle=\ttfamily\footnotesize,
  breaklines=true,
  breakindent=0pt,
  frame=single,
  caption={YAML definition for the \texttt{natural\_science} facet (biomedical domain subset).},
  label={lst:yaml}
]
modules:
  natural_science:
    type: relational
    indicators:
      P101:    # field of work
      P69:     # educated at
      P106:    # occupation
        - Q864503    # biologist
        - Q2374149   # botanist
        - Q3779582   # microbiologist
        - Q3640160   # marine biologist
        - Q350979    # zoologist
        - Q3055126   # entomologist
        - Q1225716   # ornithologist
        - Q2487799   # mycologist
        - Q15839134  # ecologist
        - Q18805     # naturalist
        - Q12773412  # parasitologist
        - Q4205432   # ichthyologist
        # ... 10 more occupations
    value_props:
      - P101   # field of work
      - P69    # educated at
      - P512   # academic degree
      - P184   # doctoral advisor
      - P185   # doctoral student
      - P1416  # affiliation
      - P803   # professorship
\end{lstlisting}

\section{LLM-Guided Extraction Prompt}\label{app:prompt}

The following system prompt is constructed directly from the schema's category taxonomy and module tag vocabulary. It is provided to the LLM alongside a text passage; the model returns structured JSON with extracted entities, their category, context-relevant tags, and an explanatory context field.

\begin{lstlisting}[
  basicstyle=\ttfamily\footnotesize,
  breaklines=true,
  breakindent=0pt,
  frame=single,
  caption={System prompt for schema-guided entity extraction.},
  label={lst:prompt},
  float=tp
]
You are an entity extraction system. Given a text passage, extract all named entities and classify each one into a category, then assign relevant tags.

### Entity Categories

Use the following taxonomy to classify each extracted entity:

| Category     | Entity Types (examples)                                        |
|--------------|----------------------------------------------------------------|
| Person       | human (real, living or historical)                             |
| Organization | business, company, university, government agency               |
| Work         | book, film, television series, fictional character             |
| Place        | street, river, human settlement, lake                          |
| Science      | planet, asteroid, chemical element, disease, gene              |
| Event        | war, battle, election, conference, natural disaster            |
| Knowledge    | scholarly article, theorem, algorithm, law                     |
| Product      | software, aircraft, vehicle, video game                        |

### Tags

For each category, assign tags from the list below if relevant to that entity under the original context -- not based on general world knowledge.

- Person: entertainment, creativity, education, politics, sports, religion, finance, military, society, crime, family, award, healthcare, legal, culture, government, technology, food
- Organization: location, affiliation, education, international, government, corporation, culture, religion, politics, sports, broadcast, finance, society, healthcare, military, legal, transportation, entertainment, award, food
- Work: literature, film, television, art, comics, periodical, revenue, museum, music, sports, military, religion, character, culture, award, authorship, location
- Place: dwelling, commercial, culture, religion, infrastructure, transportation, government, administration, sports, heritage, education, nature, finance, military, entertainment, healthcare, affiliation, architecture, food
- Science: element, compound, mineral, geology, disease, gene, biofunction, protein, pharmaceutical, anatomy, organism, botany, astronomy, clinical, healthcare, education, biosystem, clinical_trial, cell_lineage, genomics
- Event: military, conference, exhibition, disaster, violence, politics, award, sports, society, legal, entertainment, education, religion, transportation
- Knowledge: language, theorem, algorithm, specification, abstract, position, law, unit, field, phenomenon, publication, reference, name, religion, jurisdiction
- Product: cloud, aircraft, vehicle, computer, software, game, device, military, food, tool, currency, fiction, material, artifact, brand, entertainment, transportation, technology, award

### Output Format

Return a JSON object:
{
  "entities": [
    {
      "entity": "Albert Einstein",
      "category": "Person",
      "tags": ["education", "award"],
      "context": "mentioned as Nobel Prize-winning physicist at Princeton"
    }
  ]
}

### Instructions

1. Extract all named entities from the text
2. Classify each into exactly one category
3. Assign tags only when relevant in the given context
4. Include the context field to explain tag assignments
\end{lstlisting}

The prompt maps directly to the schema: the eight categories correspond to the eight classification categories, and the per-category tag lists are the relational module names from the YAML definitions. No Wikidata-specific training is required---any LLM with instruction-following capability can use this prompt to produce schema-aligned extractions.

\end{document}